\documentclass{article} 
\usepackage{nips13submit_e,times}
\usepackage{hyperref}
\usepackage{url}
\usepackage{graphicx,caption,amsmath,amssymb}
\usepackage{enumitem,booktabs}
\title{Dank Learning: Generating Memes Using Deep Neural Networks}

\author{
Abel L. Peirson V \\
Department of Physics\\
Stanford University\\
\texttt{alpv95@stanford.edu} \\
\And
E. Meltem Tolunay \\
Department of Electrical Engineering \\
Stanford University \\
\texttt{meltem.tolunay@stanford.edu}
}

\nipsfinalcopy 

\begin{document}

\maketitle

\begin{abstract}
We introduce a novel meme generation system, which given any image can produce a humorous and relevant caption. Furthermore, the system can be conditioned on not only an image but also a user-defined label relating to the meme template, giving a handle to the user on meme content. The system uses a pretrained Inception-v3 network to return an image embedding which is passed to an attention-based deep-layer LSTM model producing the caption - inspired by the widely recognized Show and Tell Model. We implement a modified beam search to encourage diversity in the captions. We evaluate the quality of our model using perplexity and human assessment on both the quality of memes generated and whether they can be differentiated from real ones. Our model produces original memes that cannot on the whole be differentiated from real ones.
\url{https://github.com/alpv95/MemeProject}
\end{abstract}

\section{Introduction}
'A meme is an idea, behavior, or style that spreads from person to person within a culture often with the aim of conveying a particular phenomenon, theme, or meaning represented by the meme.'  \cite{wiki}, \cite{mw} \\ \\
Memes are ubiquitous in today’s day and age; their language and ideologies are in constant flux. Memes come in almost every form of media, with new formats constantly evolving. Primarily they function as a medium for humor to be shared, utilizing cultural (especially subcultural) themes. However, they can also be manipulated to further political ideals, magnify echo chambers and antagonize minorities. Memes are their own form of communication and have truly captured this generation.
As AI grows in leaps and bounds it requires new and challenging tasks. The contemporary relevance of memes and the high level of understanding required to generate them motivate this project \cite{CS230}. \\ \\
We approach this task by considering only the image-with-caption class of meme, an example of which is shown in Fig.1. This reduces the problem greatly and allows for relatively simple collection of datasets. In this paper, we specifically refer to meme generation as the task of generating a humorous caption in a manner that is relevant to the initially provided image, which can be a meme template or otherwise. We apply an encoder-decoder image captioning system similar to the one outlined in \cite{showtell} \cite{tensorflow}, consisting of an CNN image embedding initial stage, followed by an LSTM RNN for language generation. We introduce and test several variants of the LSTM model and evaluate their outputs. \\ \\
Evaluation of generated meme quality is difficult to reliably automate. We evaluate and fine-tune our models using perplexity as a measure for the language modeling task, which is in high correlation with the BLEU score \cite{showtell},  and support our quantitative evaluations by utilizing human testers. Testers are asked to attempt to differentiate generated memes from real ones and/or rank generated memes on their hilarity \cite{CS230}, because at the end of the day their purpose is to be funny. 
\begin{figure}[h]

\captionsetup{width=.8\linewidth}
\begin{center}
\includegraphics[scale=0.35]{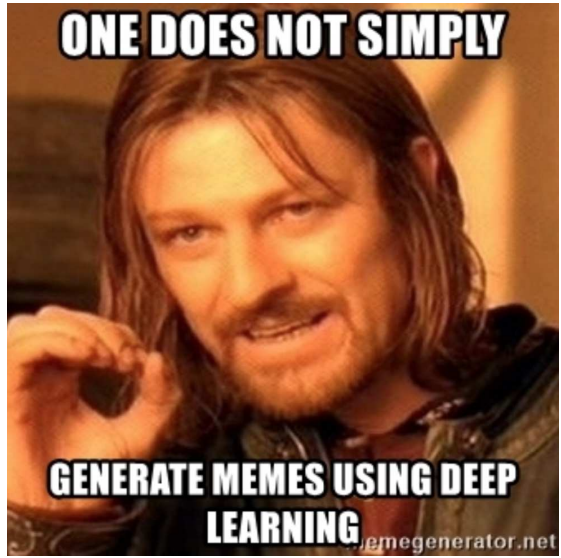}
\end{center}

\centering

\caption{\fontsize{11}{13}\selectfont A meme produced on \cite{memegenerator}, utilizing the popular Boromir format.}

\end{figure}

\section{Background/Related Work}
\subsection{Image Captioning Models}
The advent of sequence-to-sequence machine translation models \cite{sutskever_sequence_2014} established the idea of encoding information (such as a French sentence) and using at is an input to an RNN that generates language. It was not long before this encoder-decoder framework was extended to image captioning \cite{showtell}. \\ \\
Authors of \cite{showtell} introduce a widely recognized model for image captioning, which constitutes the backbone of our meme generation model. Accordingly, this model consists of an encoder-decoder scheme. The encoder is a deep convolutional neural network (CNN) that that takes images as inputs and produces fixed-length vector embeddings which are then fed into the decoder. The implementation of the decoder \cite{showtell architecture} begins with a trainable fully connected layer that takes the image embeddings from the encoder and maps them to the word embedding space. The output of this fully connected layer constructs the initial state of the RNN network which is used for language modeling, i.e. caption generation. Authors choose to use a Long Short Term Memory (LSTM) network as a variant of RNNs due to their well established success on sequence tasks \cite{showtell}. \\ \\
This type of image captioning model has been further improved recently by using bi-directional LSTMs \cite{bi} and including attention mechanisms \cite{xu_show_2015}, discussed below. Although these models perform very well on metrics such as BLEU for factual descriptions of images, there has been little work on generating humorous captions. Models such as StyleNet \cite{Stylenet} have attempted to produce humorous caption using an encoder-decoder architecture with limited success.
Successful meme generation requires a diverse range of humorous captions for the same image which are related to concepts portrayed by the image, not necessarily the content of the image itself. To achieve this we make use of much of the previous work above while incorporating our own ideas.

\subsection{Recurrent Neural Networks (RNNs) for Language Modeling Tasks}
RNNs and their variants are known to produce state-of-the-art results in sequential NLP tasks such as language modeling and machine translation. The authors of \cite{mikolov} and \cite{karpathy} discuss the success of RNNs in sequential models, where the input data does not have a fixed size. Among different types of RNNs, LSTMs \cite{lstm} are known to provide very satisfying results due to the fact that they employ "gating mechanisms" to remember data from long periods of time. The LSTM cells that we also use in our model due to same motivations operate based on the following equations \cite{showtell}:
\begin{align*}
i_t &= \sigma (W_{ix}x_{t} + W_{im}m_{t-1}) \\
f_t &= \sigma (W_{fx}x_{t} + W_{fm}m_{t-1}) \\
o_t &= \sigma (W_{ox}x_{t} + W_{om}m_{t-1}) \\
c_t &= f_t \circ c_{t-1} + i_t \circ \tanh(W_{cx}x_{t} + W_{cm}m_{t-1}) \\
m_t &= o_t \circ c_t \\
p_{t+1} &= \textrm{softmax}(m_t)
\end{align*}
where $f$ is the forget gate, $i$ the input gate, $o$ is the output gate, $m$ is the memory output and $W$ are the trainable matrices. The word prediction is made through a softmax layer which outputs a probability distribution for each word in the vocabulary.
\subsection{Pretrained GloVe Vectors}
Using vector embeddings to represent words is a vital concept to capture semantic similarities in various NLP tasks. Therefore, we rely on vector embeddings that have been previously trained on very large corpus to capture the semantic connections necessary for our text generation task. In this context, pretrained GloVe \cite{glove} word embeddings constitute the most suitable choice of word representation for our project, based on the fact that they have been trained on very large corpora, offering an option of words from Common Crawl with 42 billion tokens, 1.9 million uncased vocabulary and 300 dimensional embeddings \cite{gloveembed}. Our choice of these vector embeddings relies heavily on the words included in our meme caption dataset. Memes often include informal and slang words, and our analysis of the Common Crawl GloVe dictionary confirms that most of those words are available as pretrained GloVe embeddings in the crawl. Our incorporation of the GloVe vectors into the model will be explained in the following sections.

\subsection{Attention Mechanisms for RNNs}
Attention is one of the most revolutionary concepts introduced to deep learning with state-of-the-art results. In sequential NLP tasks such as language modeling/text generation and machine translation, attention offers a solution to the bottleneck problem which occurs from using fixed-length vectors for long input sequences \cite{bahdanau}. The general idea of attention is for the decoder to be able to pick up relevant embeddings in the encoder without running into a memory issue. Two of the most common variants of attention are introduced by Bahdanau et al. \cite{bahdanau} and Luong et al.\cite{attention}, the latter of which offers adjustments to the Bahdanau et al. attention model, and we choose to proceed with this attention mechanism in one of the variants of our model. \\ 

\section{Approach}
\subsection{Dataset}
The structure of our dataset has a significant influence on our model design and therefore shall be introduced before a thorough model description. Our dataset consists of approximately 400.000 image, label and caption triplets with 2600 unique image-label pairs, acquired from \cite{memegenerator} using a python script that we wrote. Labels are short descriptions referring to the image, i.e. the meme template, and are the same for identical images. Accordingly, each image-label pair is associated with several (roughly 160) different captions. A sample from our dataset is shown in table 1.
\begin{table}[h!]
     \begin{center}
     \begin{tabular}{ c  p{5cm}  p{5cm}  }
     \toprule
      Image & Label & Caption \\ 
    \cmidrule(r){1-1}\cmidrule(lr){2-2}\cmidrule(l){3-3}
     \raisebox{-\totalheight}{\includegraphics[scale=0.4]{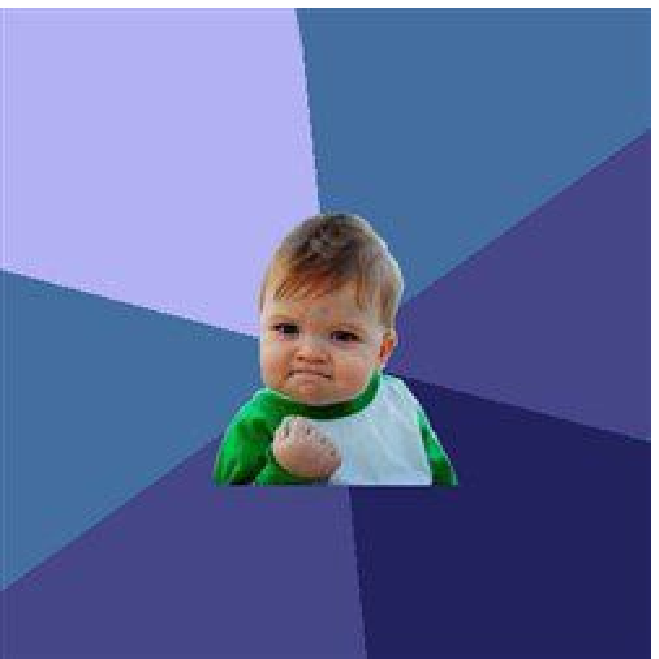}}
      & 
      \begin{itemize}[topsep=0pt]
      \item success kid
      \end{itemize}
      & 
      \begin{itemize}[topsep=0pt]
      \item Didnt study for a test still get a higher grade than someone who did
      \item Ate spaghetti with a white shirt on no stains
      \item New neighbors Free Wifi \ldots
      \end{itemize} \\
      \cmidrule(r){1-1}\cmidrule(lr){2-2}\cmidrule(l){3-3}
     \raisebox{-\totalheight}
      {\includegraphics[scale=0.4]{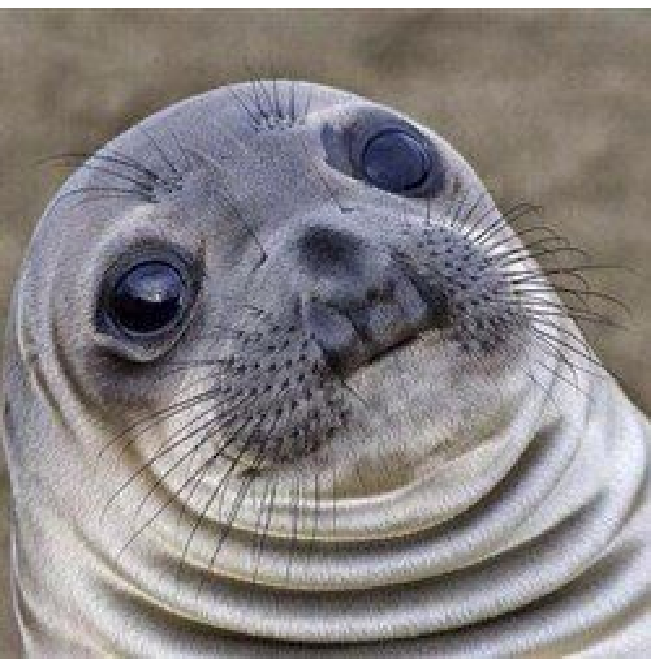}}
      & 
      \begin{itemize}[topsep=0pt]
      \item awkward seal
      \end{itemize}
      & 
      \begin{itemize}[topsep=0pt]
      \item You laugh when your friend says something He was being serious
      \item took a photo camera the wrong way 
      \item Goes to friends house Friend isn't there yet \ldots
      \end{itemize}

      \\ \bottomrule
      \end{tabular}
      \caption{Sample dataset}
      \label{tbl:myLboro}
      \end{center}
      \end{table}

Before training our model with the captions, we performed a preprocessing on our dataset. Each word in the caption was lowercased to match the GloVe format, punctuation marks were assigned their own tokens, and start, end and unknown tokens were introduced into the vocabulary with randomly initialized word vectors. We performed a cut on the words that appear in the captions: every word that appears fewer than 3 times in the corpus is set to UNK. Captions with more than 2 UNKs are removed from the dataset. 

\subsection{Model Variants}
\subsubsection{Encoder}
The motivation behind the encoder is to provide a meaningful initial state to the decoder to initiate the text generation process. To capture the image embeddings, we rely on the CNN model Inception-v3 \cite{inception-v3},\cite{showtell} pretrained on the ILSVRC-2012-CLS image classification dataset. We take the last hidden layer of the CNN as the encoder output. When a meme template is run through the Inception model, the output is a fixed length vector, image embedding, that captures the content of the image. Note that this model outputs a 2048 dimensional vector which results in a mismatch with our word embedding space that is 300 dimensional. Hence, we project the image embeddings into the word embedding space using a trainable fully connected layer. 

In our project, we implement 3 different variants of the proposed encoder scheme. The first variant explained above just uses the meme templates and disregards the labels completely. Hence, the inputs to the decoder solely include encodings from the images. Only an image is required to generate a meme. This model and its results can be seen in \cite{CS230}.

The second variant of the encoder includes the meme labels. In this model, we first obtain the image embeddings by running the images though Inception as previously. Now, we also get the pretrained GloVe embedding for each word present in the meme label and compute their average. This averaged vector is concatenated to the image embedding vector and then fed into a trainable fully connected layer.
We average rather than concatenate as this keeps size constant. This was motivated by the assumption that the label words contain semantic encodings that can be mapped into word embedding space to deliver contextual meaning for the language generation that occurs in the decoder. The output of the fully connected layer is fed into the decoder as the initial state in the same manner as the first encoder scheme. Fig. 2 shows this architecture.

The third variant of our model again includes the meme labels, but makes a slight adjustment to the encoder to include attention mechanism. In this model variant, we obtain the image embeddings and put them through a fully connected layer, but this time we extend the encoder with an additional LSTM network before the decoder. This LSTM network takes the projected image embedding as the initial state and runs the GloVe embeddings of the labels through the LSTM. We perform attention on the encoder LSTM cells using Luong attention mechanism \cite{attention}. The output of this additional LSTM network serves as the initial state of the decoder.

In light of these encoder schemes, the equations that provide the initial state for the decoder in each model variant are shown below. Let $p \in \mathbb{R}^{2048} $ be the inception output corresponding to a meme template, and let $q \in \mathbb{R}^{300}$ be the decoder initial state. Also let $e_i \in \mathbb{R}^{300}$ represent the GloVe embeddings of the label words.
\begin{align*}
&(1)\quad  q=W_1p+b_1 \\
&(2)\quad  q=W_2(p || \frac{e_1+...+e_n}{n})+b_2 \\
&(3)\quad  x_t=W_3p+b_3, \quad q=m_t 
\end{align*}
where $x_t$ and $m_t$ follow the variable naming conventions of the LSTM equations in \cite{showtell}, introduced in section 2.2 of our paper. 
\subsubsection{Decoder}
The decoder consists of a unidirectional LSTM network that operates according to the equations described in section 2.2. Every LSTM cell reuses the variables in the model. One of the modifications that we introduce to the original Show and Tell Model \cite{showtell} is the use of pretrained GloVe embeddings rather than randomly initialized word vectors. In addition, we still leave the word embedding vectors trainable based on the fact that we have a very large dataset with approximately 40k unique words and that the semantics of memes are very idiosyncratic so may not be entirely captured by the pretrained vectors. We find both qualitatively and quantitatively that allowing word vectors to be trainable improves our results.

In the previous section, we introduced 3 different variants to the encoder scheme. On the other hand, we use the same decoder construction for each of the encoders, except for the last model that incorporates an attention mechanism. Furthermore, we initially constructed each variant as a single-layered model, and then increased the number of layers to 2 and 3 to perform evaluations on the effect of model depth on the language modeling task.

\begin{figure}
\captionsetup{width=1.0\linewidth}
\begin{center}
\includegraphics[scale=0.4]{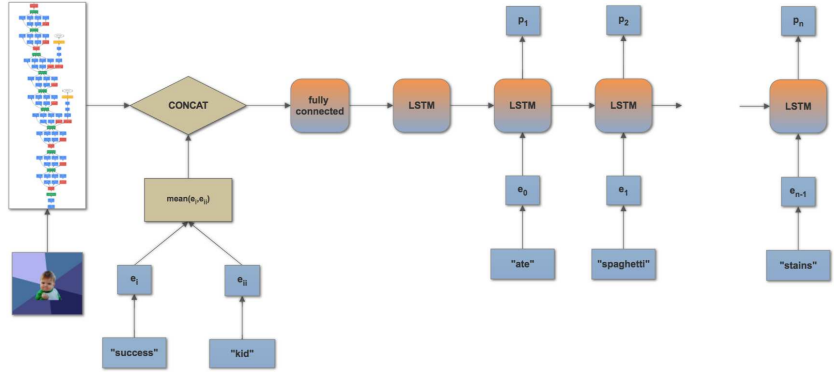}
\end{center}

\centering

\caption{\fontsize{10}{13}\selectfont A visualization of our second encoder-decoder model variant. We refer to GloVe word embeddings as $e_i$ and the probability distribution obtained at each LSTM time-step as $p_j$.}
\end{figure}

\subsubsection{Inference and Beam Search}
In the absence of label words that describe an image, inference, i.e. text generation is initiated with the image embedding. After each time-step that corresponds to a single cell, softmax probability distributions are computed for the words in the vocabulary. The output of an LSTM cell is fed sequentially into the next cell to generate the next word based on another softmax probability distribution. For original and humorous meme generation, greedy search is completely ineffective.
Furthermore, we found that inference based on standard beam search, in which we keep $k$ outputs in memory at each time-step, sequentially compute their "descendants" and then finally output the $k$ sentences with the overall highest total probability scores, gives adequate but non-optimal results.  In order to generate the freshest memes and diverse outputs for the same template we implement a temperature function into the beam search algorithm. Instead of selecting the top k most probable words, the k words are sampled from a probability distribution of the top 100 words, where the temperature of the distribution is a hyperparameter. A probability distribution $p$ can be modified with temperature $T$ by the function $f$ shown below.
\begin{equation}
f(p)_i = \frac{p_i^{1/T}}{\Sigma_jp_j^{1/T}}
\end{equation}
where $T=1$ corresponds to unchanged probabilities, high $T$ lead to a very flat distribution (random pick) and low $T$ leads to argmax (greedy search).

In the presence of label vectors that correspond to a meme template, i.e. using the second or third model variants for inference, the users can provide additional context for the text generation by providing labels of their choice, giving them an additional handle on the generated meme content.

\section{Experiments}
\subsection{Training}
For each model variant, we performed training using 1, 2 and 3 layered versions of the LSTM decoder network. In addition, we tested both Momentum and SGD optimizers. A thorough hyper-parameter search was done to find the best learning rate schedule, batch size and LSTM/attention unit size. The evaluation metric during this search was perplexity, introduced in section 4.2 and displayed for both models in fig.3. Final hyperparameter choices are also shown in fig.3  In practice, we observed no significant difference in the perplexity score or the output quality of the memes when we increased the LSTM decoder network layer depth from 1 to 2 and eventually 3.  
\begin{figure}[h!]
\captionsetup{width=1.0\linewidth}
\begin{center}
\includegraphics[scale=0.5]{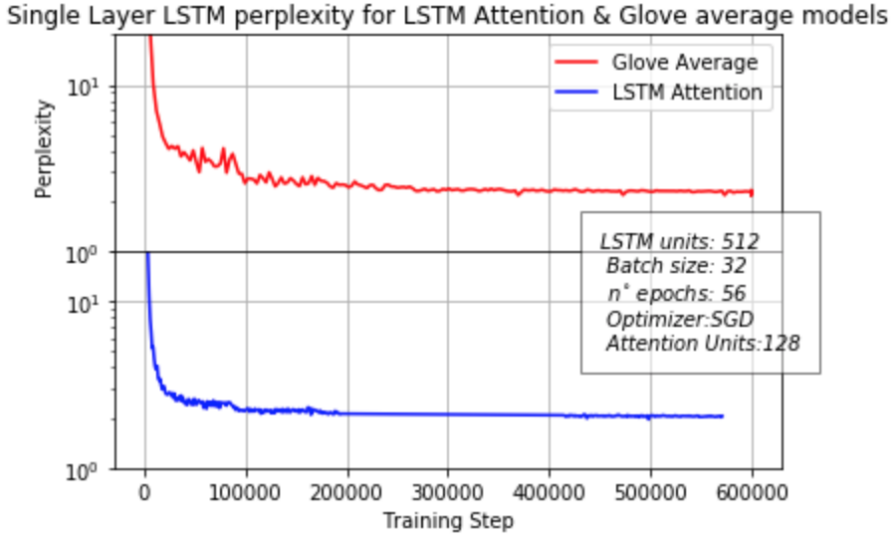}
\end{center}

\centering

\caption{\fontsize{10}{13}\selectfont A visualization of our second encoder-decoder model variant. We refer to GloVe word embeddings as $e_i$ and the probability distribution obtained at each LSTM time-step as $p_j$.}
\end{figure}

\subsection{Results and Evaluation}

For the tuning of hyperparameters and to quantitatively evaluate our model, we built an evaluation set of 105 template-label-caption examples taken from the training set which have repeated formats. E.g in Fig.1 the Boromir meme almost always starts the caption with `one does not simply', so the boromir image + the label `boromir' +`one does not simply' would constitute one example in the eval set.
We calculate the perplexity scores of the model on this eval set. Perplexity (PP) is a measure of the inverse probabilities of predicting the next word in the example caption (C):
\begin{equation}
PP(C) = \sqrt[N]{\prod_{i=1}^{N}\frac{1}{P(w_i|w_1...w_{i-1})}}.
\end{equation}
$\left\{w_1 ...w_N \right\}$ are the words in caption C. The probabilities P can be found from the cross entropy loss calculated on the LSTM output. Low perplexity means the model frequently assigns high probabilities to the given evaluation example captions. This metric tells us how well the model is learning to caption images of different formats with the correct style. It is a limited metric for success as it tells us nothing about whether the captions are humorous, original and varied. To solve this problem, we additionally implement a function to check whether generated captions, or exceptionally similar captions, are in the training set. For our final test we show 20 different memes to 5 people from diverse backgrounds and note if they can differentiate them from real (training set) memes of the same format and random text generated memes. The same people also ranked the memes being shown on how funny they found them on a scale of 0-10. The final results of these analyses are shown in fig.3 and table 2

Taking a look at our model variants' results in table 2, there are not significant differences between the two models that make use of the meme labels both in terms of perplexity and human assessments. Fig.4 shows some example generated memes from the variants, where we can see both models generalize relatively well to unseen images. The average meme produced from both is difficult to differentiate from a real meme and both variants scored close to the same hilarity rating as real memes, though this is a fairly subjective metric. The attention variant decreases the number of caption copies from the dataset compared to the GloVe average model but reduces performance on the human tested metrics. This might be expected as attention model variant places more emphasis on the labels, so by varying the labels for new images this could encourage more original memes. However there are only 2600 unique labels in the training set, so it would be difficult for the model to generalize to new labels, thus reducing performance.

Compared with the model variant that employs an image-only encoder \cite{CS230}, conditioning on the labels did allow for more varied meme content. We found that the given label did not provide a good handle on the content of the generated meme caption as can be seen in fig.4: the unseen generated memes are not related to the labels, only to the images. Again, this is to be expected considering there were few unique labels in the training set and these were only broadly related to the captions. 
\begin{figure}
\captionsetup{width=1.0\linewidth}
\begin{center}
\includegraphics[scale=0.45]{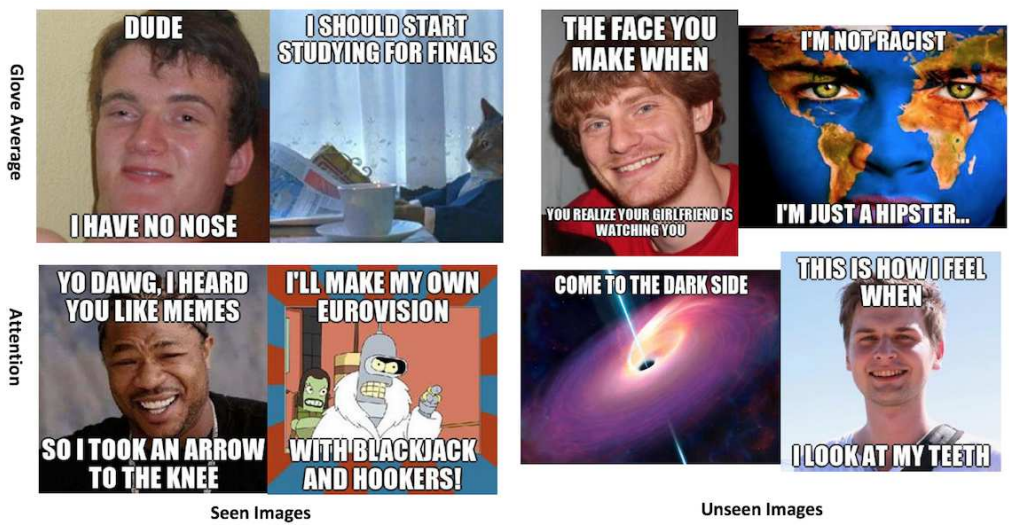}
\end{center}

\centering

\caption{\fontsize{10}{13}\selectfont Original memes generated by both model variants. The input labels for the seen images are their labels from the training set while for the unseen image we use `AI is the new electricity' for all 4.}
\end{figure}

\begin{table}[h!]
\centering
\begin{tabular}{c|c c c c} 
\hline
 Model& \% in data & Perplexity & Hilarity & Differentiability \\ 
 \hline\hline
\multicolumn{4}{c}{\textbf{Seen Images}} \\
\hline
 Attention & 16 & 2.02 & 6.0 & 70\%\\ 
 \hline
 Glove Averages & 18 & 2.28 & 6.8 & 63\% \\ 
 \hline
 \multicolumn{4}{c}{\textbf{Unseen Images}} \\
 \hline
 Attention & 18 & - & 5.5 & -\\
 \hline
 Glove Averages & 26 &  & 6.9 & -\\
\end{tabular}
\caption{Real Memes scored on average 7.0 in the hilarity test, between 1-10. Differentiability shows how often our human testers were able to distinguish real memes from generated ones, if indistinguishable this score would be near 50\%. Since memes are generated by beam search, one has to manually select a caption from the k generated, introducing a slight selection bias.}
\end{table}

\section{Conclusion}
In this paper, we have successfully demonstrated how to use a neural network model to generate memes given an input image. We suggested different variants of encoder schemes that can operate with or without labels, and reported a fine-tuned LSTM network for language modeling. Obtained results indicate that memes can be generated that in general cannot be easily distinguished from naturally produced ones, if at all, using human evaluations.

We acknowledge that one of the greatest challenges in our project and other language modeling tasks is to capture humor, which varies across people and cultures. In fact, this constitutes a research area on its own, as seen in publishings such as \cite{humor}, and accordingly new research ideas on this problem should be incorporated into the meme generation project in the future. One example would be to train on a dataset that includes the break point in the text between upper and lower for the image. These were chosen manually here and are important for the humor impact of the meme. If the model could learn the breakpoints this would be a huge improvement and could fully automate the meme generation.
Another avenue for future work would be to explore visual attention mechanisms that operate on the images and investigate their role in meme generation tasks, based on publishings such as \cite{showattendtell}, \cite{textguided} and \cite{visualattn}. 

Lastly we note that there was a bias in the dataset towards expletive, racist and sexist memes, so yet another possibility for future work would be to address this bias.

\small{

}

\end{document}